\documentclass{article}

\usepackage{arxiv}

\usepackage[utf8]{inputenc} 
\usepackage[T1]{fontenc}    
\usepackage{hyperref}       
\usepackage{url}            
\usepackage{booktabs}       
\usepackage{amsfonts}       
\usepackage{nicefrac}       
\usepackage{microtype}      
\usepackage{lipsum}

\usepackage{bm}
\usepackage{amsmath}
\usepackage{amssymb}
\usepackage{amsfonts}
\usepackage{graphicx}
\def\vec#1{\mathbf{#1}}
\usepackage{algorithm}
\usepackage{algorithmic}

\title{Meta-Active Learning for Node Response Prediction in Graphs}

\author{
  Tomoharu Iwata\\
  NTT Communication Science Laboratories\\
}
\date{}

\begin{document}
\maketitle

\begin{abstract}
  Meta-learning is an important approach to improve machine learning performance with a limited number of observations for target tasks. However, when observations are unbalancedly obtained, it is difficult to improve the performance even with meta-learning methods. In this paper, we propose an active learning method for meta-learning on node response prediction tasks in attributed graphs, where nodes to observe are selected to improve performance with as few observed nodes as possible. With the proposed method, we use models based on graph convolutional neural networks for both predicting node responses and selecting nodes, by which we can predict responses and select nodes even for graphs with unseen response variables. The response prediction model is trained by minimizing the expected test error. The node selection model is trained by maximizing the expected error reduction with reinforcement learning. We demonstrate the effectiveness of the proposed method with 11 types of road congestion prediction tasks.
\end{abstract}

\section{Introduction}

A wide variety of data are represented as graphs,
such as road networks~\cite{lammer2006scaling,geng2019spatiotemporal},
citation graphs~\cite{ding2011scientific},
metabolic pathways~\cite{kanehisa2000kegg},
ecosystem~\cite{roberts1978food}, and
social networks~\cite{wasserman1994social}.
In this paper, we consider node response prediction tasks,
where a response value for each node is predicted given an attributed graph.
Node response prediction is an important task, which includes
congestion prediction with a road network~\cite{li2017diffusion,cui2019traffic},
scientific paper classification
with a citation graph~\cite{yang2016revisiting}, and
user preference prediction with a social network~\cite{jamali2010matrix,walter2008model}.

As the number of nodes with observed responses increases,
the performance of node response prediction is improved in general.
However, a sufficient number of observed responses are often unavailable
since obtaining responses requires high cost,
e.g., placing sensors at many roads,
and manually labeling by experts with domain knowledge.
For improving performance with a small number of observations,
many meta-learning methods have been proposed~\cite{schmidhuber:1987:srl,bengio1991learning,ravi2016optimization,andrychowicz2016learning,vinyals2016matching,snell2017prototypical,bartunov2018few,finn2017model,li2017meta,kimbayesian,finn2018probabilistic,rusu2018meta,yao2019hierarchically,edwards2016towards,garnelo2018conditional,kim2019attentive,hewitt2018variational,bornschein2017variational,reed2017few,rezende2016one,tang2019,narwariya2020meta,xie2019meta,lake2019compositional}.
Meta-learning is to learn a model that can predict
unseen response variables with only a few observed nodes.
However, when observed nodes are unbalancedly placed in a graph,
e.g., observations are obtained with nodes that are directly
connected to each other and no observations are given with other distant nodes, 
it is difficult to improve the prediction performance even with meta-learning methods.

\begin{figure}[t!]
  \centering
  {\tabcolsep=2em
  \begin{tabular}{cc}
  \includegraphics[height=24em]{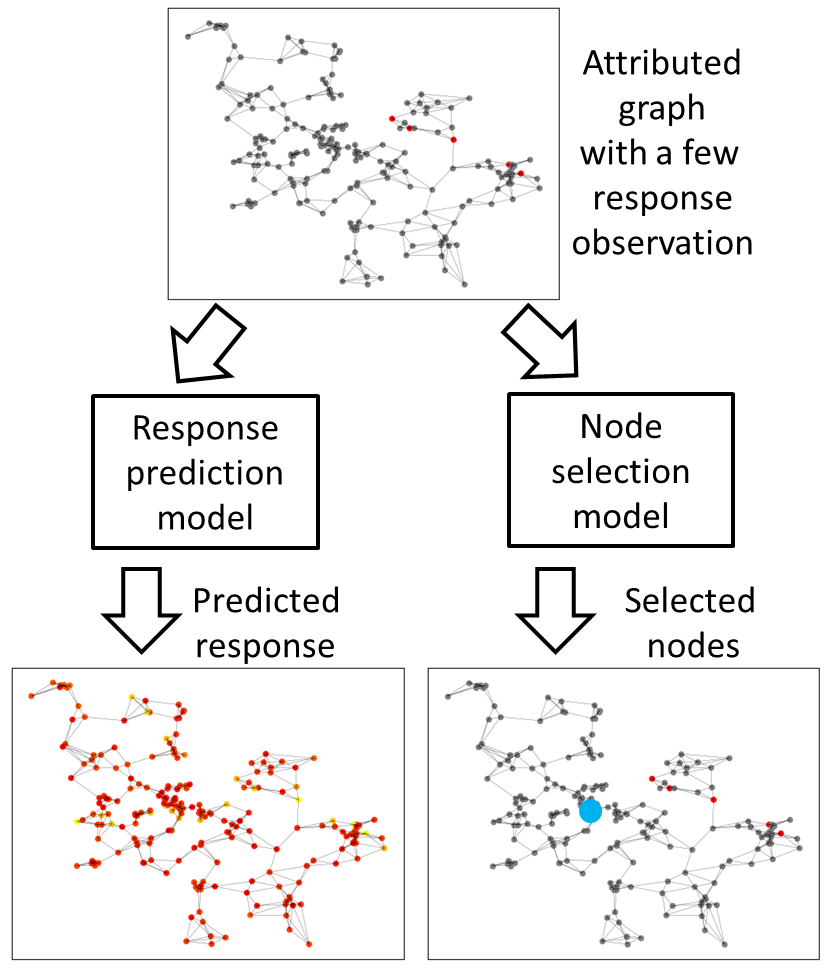}&
  \includegraphics[height=24em]{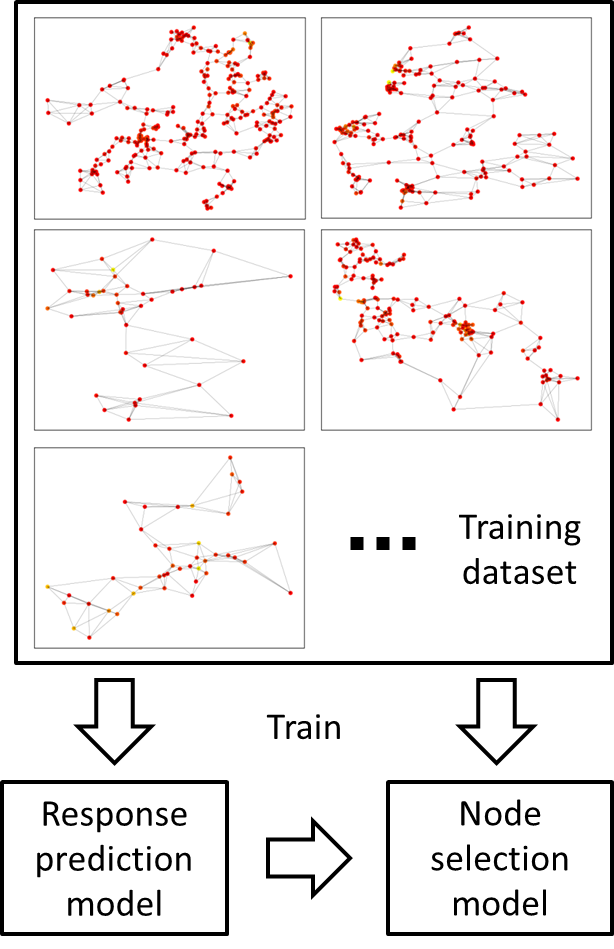}\\
  (a) Models & (b) Training \\
  \end{tabular}}
  \caption{(a) Our response prediction and node selection models. These models take attributed graphs with a few response observations as input, where observed and unobserved nodes are represented by red and gray circles. The response prediction model outputs predicted responses for each node, where red (yellow) indicates high (low) predicted values. The node selection model outputs scores for each node that are used for selecting nodes to observe in active learning, where a selected node is represented by the big blue circle. (b) Our training framework. First, the response prediction model is trained using a training dataset, which contains various graphs with various response types. Second, the node selection model is trained using the training dataset and trained response prediction model.}
  \label{fig:task}  
\end{figure}

In this paper, we propose an active learning method
to select nodes to observe in an attributed graph
so that the prediction performance is improved with as few observed nodes as possible.
We assume that many attributed graphs with observed responses are given as a training dataset.
Our task is to improve the node response prediction performance
with fewer observations in an unseen attributed graph
with an unseen response variable.
The proposed method contains two models: a response prediction model,
and node selection model.
The response prediction model predicts node responses
given an attributed graph,
and the node selection model outputs scores
for selecting a node to observe given an attributed graph.
Figure~\ref{fig:task} illustrates the response prediction and node selection models.
We use graph convolutional neural networks (GCNs)~\cite{kipf2016semi}
for both of the models.
By taking attributes, observed responses, and masks that indicating observed nodes as input,
the GCNs can output predictions or scores for unseen graphs depending on the observed responses
by aggregating information of all nodes considering the graph structure.

The response prediction model is trained by
minimizing the expected test prediction error
using an episodic training framework~\cite{ravi2016optimization,santoro2016meta,snell2017prototypical,finn2017model,li2017meta}.
The episodic training framework,
which is often used for meta-learning, simulates a test phase
by randomly selecting observed and unobserved nodes
using training attributed graphs.
The node selection model is trained by
maximizing the expected test error reduction
by reinforcement learning.
Although many existing active learning methods
use heuristics,
such as uncertainty~\cite{lewis1994sequential,holub2008entropy,yu2010active,jing2004entropy,gal2017deep},
for selecting node policies,
our meta-learning framework directly optimizes an active learning policy
that maximizes the expected test error reduction,
where the policy is applicable to unseen graphs with unseen response variables.
Figure~\ref{fig:task} illustrates our training framework.

Our main contributions are as follows:
\begin{itemize}
\item We propose a meta-active learning method for node response prediction in graphs, where our models can be used for unseen graphs with unseen response variables.
\item The proposed method directly maximizes the expected test error reduction for node response prediction tasks.
\item We demonstrate the effectiveness of the proposed method using 11 types of road congestion prediction tasks.
\end{itemize}

The remainder of this paper is organized as follows.
Section~\ref{sec:related} briefly reviews related work.
Section~\ref{sec:proposed} defines our problem,
proposes our response prediction and node selection models,
and presents their training procedures.
In Section~\ref{sec:experiments},
we demonstrate the effectiveness of the proposed method
with congestion prediction tasks using road graphs in the UK.
Finally, we give a concluding remark and future work
in Section~\ref{sec:conclusion}.

\section{Related work}
\label{sec:related}

Active learning selects examples to be labeled
for improving the performance while reducing costly labeling effort.
Many active learning methods have been proposed~\cite{fu2013survey},
such
uncertainty sampling~\cite{lewis1994sequential,holub2008entropy,yu2010active,jing2004entropy,gal2017deep},
query-by-committee~\cite{seung1992query,freund1997selective},
mutual information~\cite{guestrin2005near,houlsby2011bayesian,iwata2013active},
core-set~\cite{sener2017active},
and
mean standard deviation~\cite{kendall2015bayesian,kampffmeyer2016semantic}.
Although the metrics used in these methods are computed efficiently,
they are different from the expected test error that we want to minimize.
Active learning methods that directly reduce the test error
have been proposed~\cite{cohn1996active,roy2001toward}.
These methods use simple models that can calculate the test error
in closed form~\cite{cohn1996active},
or use sampling to estimate the test error~\cite{roy2001toward}.
On the other hand, the proposed method
directly optimizes a policy that maximizes the test error reduction,
by which we do not need to calculate the test error in a test phase.
Although a number of active learning methods using reinforcement learning have been proposed~\cite{fang2017learning,liu2018learning,liu2018translation,haussmann2019deep,konyushkova2017learning,pang2018meta,ebert2012ralf,hsu2015active,bachman2017learning,woodward2017active}, but they are not for graphs.
Active learning for graph embedding has been proposed~\cite{cai2017active}
but it is not for node response prediction tasks.
GCNs have been in a wide variety of applications~\cite{scarselli2008graph,kipf2016semi,defferrard2016convolutional,duvenaud2015convolutional,hamilton2017inductive}, including meta-learning~\cite{bose2019meta,garcia2017few}.
However, they are not used for active learning.

\section{Proposed method}
\label{sec:proposed}

\subsection{Problem definition}

In a training phase,
we are given a set of $D$ graphs with attributes and responses,
$\mathcal{G}=\{\vec{G}_{d}\}_{d=1}^{D}$,
where $\vec{G}_{d}=(\vec{A}_{d},\vec{X}_{d},\vec{y}_{d})$ is the $d$th graph,
$\vec{A}_{d}\in\{0,1\}^{N_{d}\times N_{d}}$ is the adjacency matrix,
$N_{d}$ is the number of nodes,
$\vec{X}_{d}=(\vec{x}_{dn})_{n=1}^{N_{d}}\in\mathbb{R}^{N_{d}\times J_{d}}$,
$\vec{y}_{d}=(y_{dn})_{n=1}^{N_{d}}\in\mathbb{R}^{N_{d}}$,
$\vec{x}_{dn}\in\mathbb{R}^{J_{d}}$ is the attributes of the $n$th node,
and $y_{dn}\in\mathbb{R}$ is its response.
Although we assume undirected and unweighted graphs for simplicity,
we can straightforwardly extend the proposed method for directed and/or weighted graphs.
In a test phase,
we are given target graph 
$\vec{G}_{*}=(\vec{A}_{*},\vec{X}_{*})$ without responses.
The response type of the target graph is different from
that of the training graphs, e.g.,
the target response variable is bicycle traffic,
and the training response variables are
car, taxis and bus traffic.
Our task is to improve the response prediction performance
of all nodes in the target graph by selecting nodes to observe,
where a smaller number of observed nodes is preferred.

\subsection{Model}

Graph convolutional neural networks (GCNs)
are used for modeling both predicting responses and
selecting nodes to observe.
Let $\vec{m}_{d}\in\{0,1\}^{N_{d}}$ be the binary mask
vector for indicating observed nodes,
where $m_{dn}=1$ if the response of the $n$th node is observed,
and $m_{dn}=0$ otherwise.
Let $\bar{\vec{y}}_{d}$ be the observed response vector,
where $\bar{y}_{dn}=y_{dn}$ if $m_{dn}=1$, $\bar{y}_{dn}=0$ otherwise.
For the input of a GCN, we use the following concatenated vector of the 
attributes, observed responses, and masks,
\begin{align}
  \vec{z}_{dn}^{(0)}=[\vec{x}_{dn},\bar{y}_{dn},m_{dn}],
  \label{eq:input}
\end{align}
where $\vec{z}_{dn}^{(0)}$ is the input vector of the $n$th node, and
$[\cdot,\cdot]$ represents a concatenation.
With the above input,
we can output predictions and scores without retraining
even when responses of nodes are additionally observed
by changing the input such that
$\bar{y}_{dn}=y_{dn}$ and $m_{dn}=1$ for additionally observed node $n$.

With GCNs, the hidden state at the next layer is calculated by
\begin{align}
  \vec{z}_{dn}^{(k+1)}=\sigma\left(
  \vec{W}^{(k)}\vec{z}_{dn}^{(k)}+
  \frac{1}{N_{dn}}\sum_{m=1}^{N_{d}}a_{dnm}
  \vec{U}^{(k)}\vec{z}_{dm}^{(k)}
  \right),
  \label{eq:gcn}
\end{align}
where
$\vec{z}_{dn}^{(k)}$ is the hidden state of the $n$th node at the $k$th layer,
$\sigma$ is the activation function,
$a_{dnm}$ is the $(n,m)$th element of adjacency matrix $\vec{A}_{d}$,
$N_{dn}=\sum_{m}a_{dnm}$ is the number of neighbors of the $n$th node,
$\vec{W}^{(k)}\in\mathbb{R}^{H_{k+1}\times H_{k}}$ and
$\vec{U}^{(k)}\in\mathbb{R}^{H_{k+1}\times H_{k}}$ are the linear
projection matrices of the $k$th layer, and
$H_{k}$ is the hidden state size of the $k$th layer.
Eq.~(\ref{eq:gcn}) aggregates the information of the own node (the first term) and
its neighbor nodes (the second term) with transformation.
The hidden state at the last $K$th layer is the output of the GCN.

We use a GCN for response prediction model $f$
with parameters $\bm{\Phi}=\{\vec{W}_{\mathrm{f}}^{(k)},\vec{U}_{\mathrm{f}}^{(k)}\}_{k=1}^{K_{\mathrm{f}}}$
as follows,
\begin{align}
  \hat{\vec{y}}_{d}=f(\vec{G}_{d},\vec{m}_{d};\bm{\Phi}),
  \label{eq:f}  
\end{align}
where $\hat{\vec{y}}_{d}\in\mathbb{R}^{N_{d}}$ is the predicted responses,
and the $K_{\mathrm{f}}$ is the number of layers.
The input of the GCN in Eq.~(\ref{eq:input}) is calculated using $\vec{G}_{d}$ and $\vec{m}_{d}$.
Also, we use another GCN for node selection model $g$
with parameters $\bm{\Theta}=\{\vec{W}_{\mathrm{g}}^{(k)},\vec{U}_{\mathrm{g}}^{(k)}\}_{k=1}^{K_{\mathrm{g}}}$
as follows,
\begin{align}
  \vec{s}_{d}=g(\vec{G}_{d},\vec{m}_{d};\bm{\Theta}),
  \label{eq:g}
\end{align}
where $\vec{s}_{d}=(s_{dn})_{n=1}^{N_{d}}\in\mathbb{R}^{N_{d}}$, and
$s_{dn}$ is the score that the $n$th node is selected to observe
the response.

Our model can be used for meta-learning,
where values of unseen response variables are predicted,
since it has similar operations to existing meta-learning methods,
such as conditional  neural processes~\cite{garnelo2018conditional}.
Conditional neural processes
consist of an encoder, an aggregator, and a decoder.
The encoder takes observed attributes and responses as input,
and outputs a representation for each observed node.
The aggregator summarizes the set of representations for the observed nodes
into a single representation.
The decoder takes the aggregated representation
and attributes without responses as input,
and predicts responses.
Since our model takes the attributes, observed responses, and masks as input,
the GCN simultaneously works as an encoder, aggregator, and decoder.
In particular, the first term in Eq.~(\ref{eq:gcn}) works as an encoder
for nodes with observed responses, and 
as a decoder for nodes without observed responses.
The second term in Eq.~(\ref{eq:gcn}) works as an aggregator
by summarizing representations of the neighbor nodes.

\subsection{Training}

First, we train parameters $\bm{\Phi}$ of response prediction model $f$,
and then train parameters $\bm{\Theta}$ of node selection model $g$ while fixing $f$.

\subsubsection{Response prediction model}

We estimate parameters $\bm{\Phi}$ of
response prediction model $f$ in Eq.~(\ref{eq:f})
by minimizing the expected test prediction error,
\begin{align}
  \hat{\bm{\Phi}}=\arg\min_{\bm{\Phi}}\mathbb{E}_{\vec{G}_{d}}[\mathbb{E}_{\vec{m}_{d}}[L(\vec{G}_{d},\vec{m}_{d};\bm{\Phi})]],
\end{align}
using training graph set $\mathcal{G}$
with an episodic training framework.
Here, $\mathbb{E}$ represents an expectation,
and the expectation are taken over various graphs $\vec{G}_{d}$ in training graph set $\mathcal{G}$,
and over various observation patterns $\vec{m}_{d}$ in each graph,
where we randomly generate target graphs for simulating a test phase.
The test prediction error for a target graph is calculated by
\begin{align}
  L(\vec{G}_{d},\vec{m}_{d};\bm{\Phi})=
  \frac{1}{\sum_{n=1}^{N_{d}}(1-m_{dn})}
  \sum_{n=1}^{N_{d}}(1-m_{dn})\parallel y_{dn}-f_{n}(\vec{G}_{d},\vec{m}_{d};\bm{\Phi})\parallel^{2},
  \label{eq:L}
\end{align}
where $f_{n}(\vec{G}_{d},\vec{m}_{d};\bm{\Phi})$
is the $n$th element of the output of the response prediction model,
the responses are predicted using the observed nodes, $m_{dn}=1$,
and the test prediction error is calculated for the unobserved nodes, $m_{dn}=0$.
When response variables are categorical,
the cross-entropy loss with the softmax function can be used instead of
the squared error loss.

Algorithm~\ref{alg:prediction} shows the training procedures of the response
prediction model.
For each epoch, we simulate a test phase by uniform randomly sampling a graph
from the training dataset (Line~3) and uniform randomly selecting observed nodes
in the graph (Line~4).

\begin{algorithm}[t!]
  \caption{Training procedure of response prediction model $f$.}
  \label{alg:prediction}
  \begin{algorithmic}[1]
    \renewcommand{\algorithmicrequire}{\textbf{Input:}}
    \renewcommand{\algorithmicensure}{\textbf{Output:}}
    \REQUIRE{Set of training graphs $\mathcal{G}$,
      support set size $N_{\mathrm{S}}$}
    \ENSURE{Trained model parameters $\bm{\Phi}$}
    \STATE Initialize parameters $\bm{\Phi}$ randomly
    \WHILE{not done}
    \STATE Sample graph $\vec{G}_{d}$ from $\mathcal{G}$
    \STATE Sample $N_{\mathrm{S}}$ observed nodes $\mathcal{S}$ from $\{1,\cdots,N_{d}\}$
    \STATE Set mask vector $\vec{m}_{d}$ using the sampled observed nodes, such that $m_{dn}=1$ if $n\in\mathcal{S}$, and $m_{dn}=0$ otherwise
    \STATE Calculate loss $L(\vec{G}_{d},\vec{m}_{d};\bm{\Phi})$ by Eq.~(\ref{eq:L}) and its gradients 
    \STATE Update parameters $\bm{\Phi}$ using the gradients
    \ENDWHILE
  \end{algorithmic}
\end{algorithm}

\subsubsection{Node selection model}

We estimate parameters $\bm{\Theta}$ of node selection model $g$ in Eq.~(\ref{eq:g})
by maximizing the expected test error reduction 
using training graph set $\mathcal{G}$
based on reinforcement learning with an episodic training framework.
For rewards, we use the test error reduction
when node $n$ is selected to observe with graph $\vec{G}_{d}$ and mask $\vec{m}_{d}$
as follows,
\begin{align}
  R(\vec{G}_{d},\vec{m}_{d},n)=
   \frac{L(\vec{G}_{d},\vec{m}_{d};\hat{\bm{\Phi}})-L(\vec{G}_{d},\vec{m}_{d}^{(+n)};\hat{\bm{\Phi}})}{L(\vec{G}_{d},\vec{m}_{d};\hat{\bm{\Phi}})},
  \label{eq:r}
\end{align}
where
$\vec{m}_{d}^{(+n)}$ is the updated mask vector of $\vec{m}_{d}$
when node $n$ is additionally observed, $m_{dn'}^{(+n)}=1$ if $n'=n$,
and $m_{dn'}^{(+n)}=m_{dn'}$ otherwise.
Here, parameters of trained response prediction models $\hat{\bm{\Phi}}$ is used.
We can calculate the error
$L(\vec{G}_{d},\vec{m}_{d}^{(+n)};\hat{\bm{\Phi}})$ when node $n$ is additionally observed
by feeding the updated mask vector and graph
into our response prediction model based on GCNs
without retraining. 
In terms of reinforcement learning,
a pair of graph $\vec{G}_{d}$ and mask $\vec{m}_{d}$ is a state,
node to observe $n$ is an action,
and node selection model $g(\vec{G}_{d},\vec{m}_{d};\bm{\Theta})$ that outputs scores of actions
given a state is a policy.
The expected test error reduction is calculated by
\begin{align}
  \hat{\bm{\Theta}}=\arg\max_{\bm{\Theta}}
  \mathbb{E}_{\vec{G}_{d}}[\mathbb{E}_{(\vec{m},n)\sim\bm{\pi}(\bm{\Theta})}[R(\vec{G}_{d},\vec{m},n)]],
\end{align}
where $\bm{\pi}(\bm{\Theta})$ is the probability distribution of the policy for selecting nodes,
which is defined by node selection model $g$ with parameter $\bm{\Theta}$.

Algorithm~\ref{alg:active} shows the training procedures
of the node selection model with policy gradients,
where active learning is simulated using randomly selected graphs (Line~3).
Each active learning task starts with a graph without observed responses (Line~4),
and we iterate until $N_{\mathrm{A}}$ nodes are observed (Line~5).
A node is selected according to the following policy that is calculated
from scores $\vec{s}_{d}$ (Lines~6--7),
\begin{align}
  \pi_{dn}=\frac{\exp(s'_{dn})}{\sum_{m=1}^{N_{d}}\exp(s'_{dm})},
  \label{eq:pi}
\end{align}
where $s'_{dn}=s_{dn}$ if $m_{dn}=0$, and $s'_{dn}=-\infty$ otherwise,
by which already observed nodes are excluded from selection.
A node is sampled according to the categorical distribution with parameters $\bm{\pi}_{d}$ (Line~8).
The test error reduction, or reward, is calculated at Line~9.
At Line~10, parameters $\bm{\Phi}$ are updated by using the log-derivative trick,
or Reinforce~\cite{williams1992simple},
where the average total reward was used for the baseline~\cite{zhao2011analysis}.
Although we use the myopic rewards in the algorithm,
we can also use non-myopic rewards.

\begin{algorithm}[t!]
  \caption{Training procedure of node selection model $g$.}
  \label{alg:active}
  \begin{algorithmic}[1]
    \renewcommand{\algorithmicrequire}{\textbf{Input:}}
    \renewcommand{\algorithmicensure}{\textbf{Output:}}
    \REQUIRE{Set of training graphs $\mathcal{G}$,
      maximum support set size $N_{\mathrm{A}}$}
    \ENSURE{Trained model parameters $\bm{\Theta}$}
    \STATE Initialize parameters $\bm{\Theta}$ randomly
    \WHILE{not done}
    \STATE Sample graph $\vec{G}_{d}$ from $\mathcal{G}$
    \STATE Initialize mask vector $\vec{m}_{d}=\bm{0}$
    \FOR{$t=1,\cdots,N_{\mathrm{A}}$}
    \STATE Calculate score $\vec{s}_{d}=g(\vec{G}_{d},\vec{m}_{d};\bm{\Theta})$
    \STATE Set policy $\bm{\pi}_{d}$ by Eq.~(\ref{eq:pi})
    \STATE Sample an observed node according to the policy
    $n\sim\mathrm{Categorical}(\bm{\pi}_{d})$
    \STATE Calculate test error reduction $R(\vec{G}_{d},\vec{m}_{d},n)$ by Eq.~(\ref{eq:r})
    \STATE Update parameters $\bm{\Theta}\leftarrow\bm{\Theta}+\alpha R(\vec{G}_{d},\vec{m}_{d},n) \nabla_{\bm{\Theta}}\log \pi_{dn}$
    \STATE Update mask to include selected node $n$ in observed nodes $m_{dn}=1$
    \ENDFOR
    \ENDWHILE
  \end{algorithmic}
\end{algorithm}

\subsection{Test}

In a test phase, given target $\vec{G}_{*}$ without responses,
we first initialize mask vector $\vec{m}_{*}=\bm{0}$.
Second, we
select a node with the maximum score
\begin{align}
  \hat{n}=\arg\max_{n:m_{*n}=0}s_{*n},
\end{align}
among unobserved nodes
using the trained node selection model.
Third, we update the mask vector with the selected node by $m_{*\hat{n}}=1$.
We iterate the second and third steps until an end condition is satisfied.

\section{Experiments}
\label{sec:experiments}

\subsection{Data}

We evaluated the proposed method with
11 types of congestion prediction tasks using road graphs in the UK.
The original data were obtained 
from UK Department for Transport~\footnote{\url{https://roadtraffic.dft.gov.uk/downloads}}.
The data contained road level Annual Average Daily Flow (AADF)
at major and minor roads in the UK
for the following 11 types:
pedal cycles,
two-wheeled motor vehicles,
car and taxis,
buses and coaches,
light goods vehicles (LGVs),
two-rigid axle heavy good vehicle (HGVs),
three-rigid axle HGVs,
four or more rigid axle HGVs,
three or four-articulated axle HGVs,
five-articulated axle HGVs, and
six-articulated axle HGVs.
We used the 11 AADFs for response variables,
and used road categories, road types, longitude and latitude
for attributes.
There were six road categories, and two road types (major and minor).
These categorical attributes were transformed to one-hot vectors.
The real-valued attributes and responses were normalized in the range of zero to one.
We generated an undirected road graph for each local authority,
where a node was a road, and nodes with four nearest neighbors based on
their latitude and longitude were connected with edges.
Examples of the generated road graphs
are shown in Figure~\ref{fig:task}.
There were 206 graphs in total for each response variable and for each year,
where the average, minimum and maximum number of nodes for each graph was 103, 5 and 562.
For each experiment,
we used an AADF in a region for target data,
and used the other AADFs in the other regions for training and validation data; i.e.,
the AADF or region in target data are not contained in training and validation data.
There were 11 regions,
such as East-Midlands, London, Scotland and Wales.
A region was used for validation, and the remaining
nine regions were used for training.


\subsection{Comparing methods}

We compared the proposed active learning method with the following methods for selecting nodes to observe:
Random, Variance, Entropy, MI (mutual information), and FF (feed-forward neural network).
All the methods used the same response prediction model in Eq.~(\ref{eq:f})
to evaluate the active learning performance.
The Random method randomly selects a node to observe.

The Entropy method selects a node that maximizes the entropy, or uncertainty~\cite{lewis1994sequential,holub2008entropy,yu2010active,jing2004entropy,gal2017deep}.
With the entropy method, first, we trained a GCN that outputs mean and standard deviation of responses,
\begin{align}
  \hat{\vec{y}}_{d},\hat{\bm{\sigma}}_{d} = f_{\mathrm{entropy}}(\vec{G}_{d},\vec{m}_{d};\bm{\Phi}),
  \label{eq:f_entropy}
\end{align}
by minimizing the negative Gaussian likelihood loss instead of the test prediction error in Eq.~(\ref{eq:L}),
\begin{align}
  L_{\mathrm{entropy}}(\vec{G}_{d},\vec{m}_{d};\bm{\Phi})=
  \frac{1}{\sum_{n=1}^{N_{d}}(1-m_{dn})}
  \sum_{n=1}^{N_{d}}(1-m_{dn})\log \mathcal{N}(y_{dn};\hat{y}_{dn},\hat{\sigma}_{dn}^{2}),
  \label{eq:L_entropy}
\end{align}
where $\mathcal{N}(x;\mu,\sigma^{2})$ is the probability density function at $x$
of the Gaussian distribution with mean $\mu$ and standard deviation $\sigma$,
and $\hat{y}_{dn}$ and $\hat{\sigma}_{dn}$ are the estimated mean and standard deviation for the $n$th node by Eq.~(\ref{eq:f_entropy}).
The GCN in Eq.~(\ref{eq:f_entropy}) is the same with Eq.~(\ref{eq:f}) except that
its final layer additionally outputs standard deviation.
Using the trained GCN in Eq.~(\ref{eq:f_entropy}),
we select a node that maximizes the entropy,
$\hat{n}=\arg\max_{n:m_{*n}} \mathbb{H}[y_{dn}|\vec{G}_{d},\vec{m}_{d}]$,
where
the entropy is calculated by
\begin{align}
  \mathbb{H}[y_{dn}|\vec{G}_{d},\vec{m}_{d}]=\log\hat{\sigma}_{dn}+\frac{1}{2}(\log(2\pi)+1).
  \label{eq:H}
\end{align}

The Variance method selects a node that maximizes the variance estimated using dropout~\cite{tsymbalov2018}.
Dropout randomly sets hidden unit activities to zero~\cite{srivastava2014dropout}.
Let $\hat{y}_{dn}^{(\ell)}\in\mathbb{R}^{N_{d}}$ be the $\ell$th output of the GCN in Eq.~(\ref{eq:f})
in stochastic runs with dropout.
The variance of prediction $\hat{y}_{dn}$ was calculated by 
\begin{align}
  \mathrm{Var}[\hat{y}_{dn}] = \frac{1}{L}\sum_{\ell=1}^{L}(\hat{y}_{dn}^{(\ell)}-\bar{\hat{y}}_{dn})^{2},
\end{align}
where $L$ is the number of stochastic runs,
$\bar{\hat{y}}_{dn}=\frac{1}{L}\sum_{\ell=1}^{L}\hat{y}_{dn}^{(\ell)}$ is the average over the stochastic runs.
We used $L=10$.

The MI method selects a node that maximizes the mutual infromation~\cite{gal2017deep} between responses and model parameters,
\begin{align}
  \mathbb{I}[y_{dn},\bm{\Phi}|\vec{G}_{d},\vec{m}_{d}] = \mathbb{H}[y_{dn}|\vec{G}_{d},\vec{m}_{d}]-\mathbb{E}_{p(\bm{\Phi}|\vec{G}_{d},\vec{m}_{d})}[\mathbb{H}[y_{dn}|\vec{G}_{d},\vec{m}_{d},\bm{\Phi}]].
\end{align}
The first term was calculated with Eq.~(\ref{eq:H}).
The second term was calculated using dropout as follows,
\begin{align}
  \mathbb{E}_{p(\bm{\Phi}|\vec{G}_{d},\vec{m}_{d})}[\mathbb{H}[y_{dn}|\vec{G}_{d},\vec{m}_{d},\bm{\Phi}]]=\frac{1}{L}\sum_{\ell=1}^{L}(\log\hat{\sigma}_{dn}^{(\ell)}+\frac{1}{2}(\log(2\pi)+1))),
\end{align}
where $\hat{\sigma}_{dn}^{(\ell)}$ is the $\ell$th estimate of the standard deviation by Eq.~(\ref{eq:f_entropy})
in stochastic runs with dropout.

The FF method used feed-forward neural networks for the node selection model instead of
Eq.~(\ref{eq:g}) based on GCNs.
The FF method takes attributes $\vec{x}_{dn}$, and outputs score $s_{dn}$ for each node.
The neural networks were trained in the same way with the proposed method.
Since the FF method outputs the score for each node individually,
it does not meta-learn, but learns the relationship between the attributes and scores.

\subsection{Proposed method setting}

With GCNs for response prediction and node selection models,
the number of hidden units was 32,
the number of layers was three with residual connections.
The models were trained by
Adam~\cite{kingma2014adam} with learning rate $10^{-3}$ and dropout rate $0.1$.
The maximum number of epochs was 1,000,
and the validation data were used for early stopping.
The support set size was $N_{\mathrm{S}}=5$ for training response prediction models,
and the maximum support set size was $N_{\mathrm{A}}=10$ for training node selection models. 

\subsection{Results}

\begin{figure}[t!]
  \centering
  {\tabcolsep=0.0em
  \begin{tabular}{ccc}
  \includegraphics[height=11em]{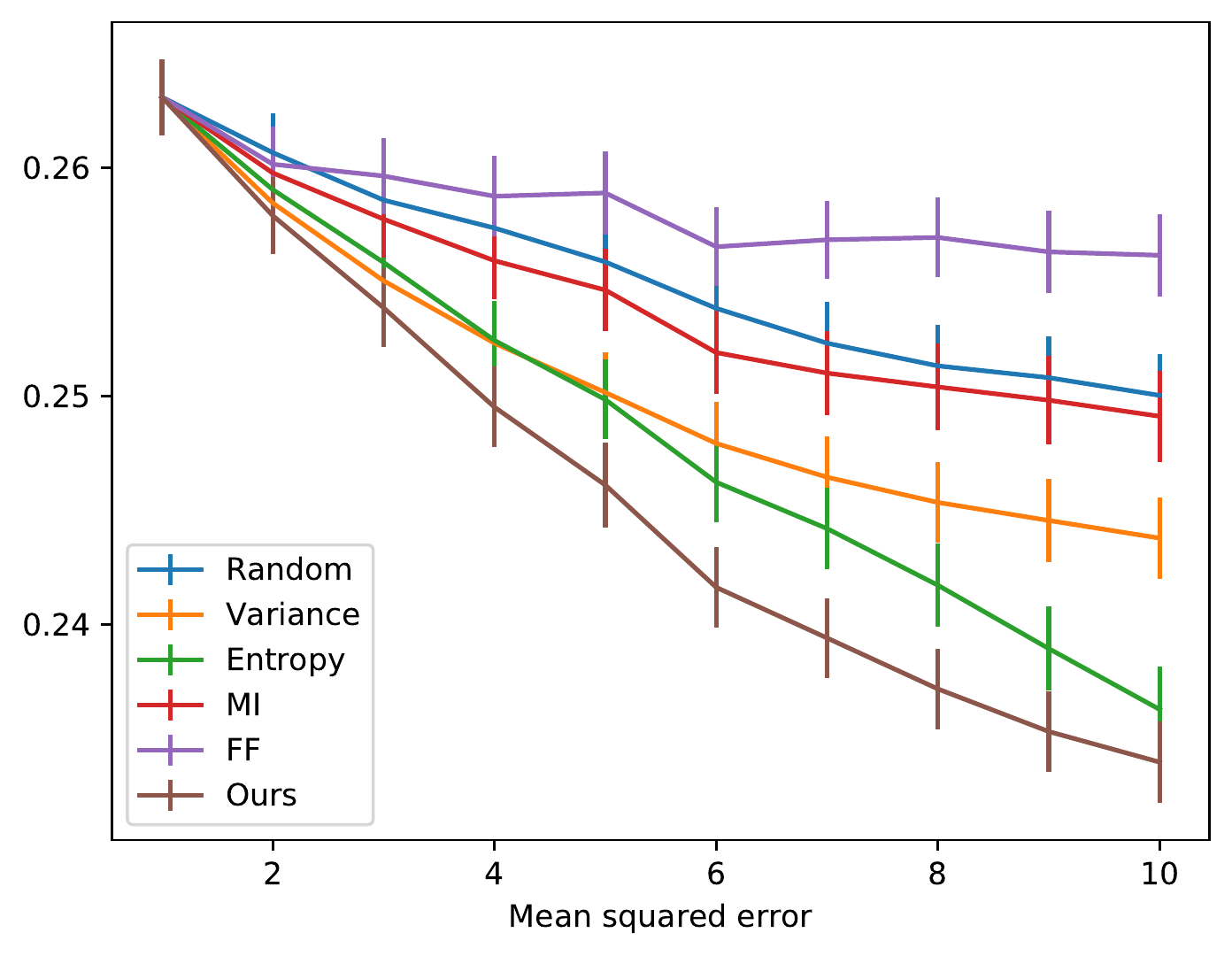}&
  \includegraphics[height=11em]{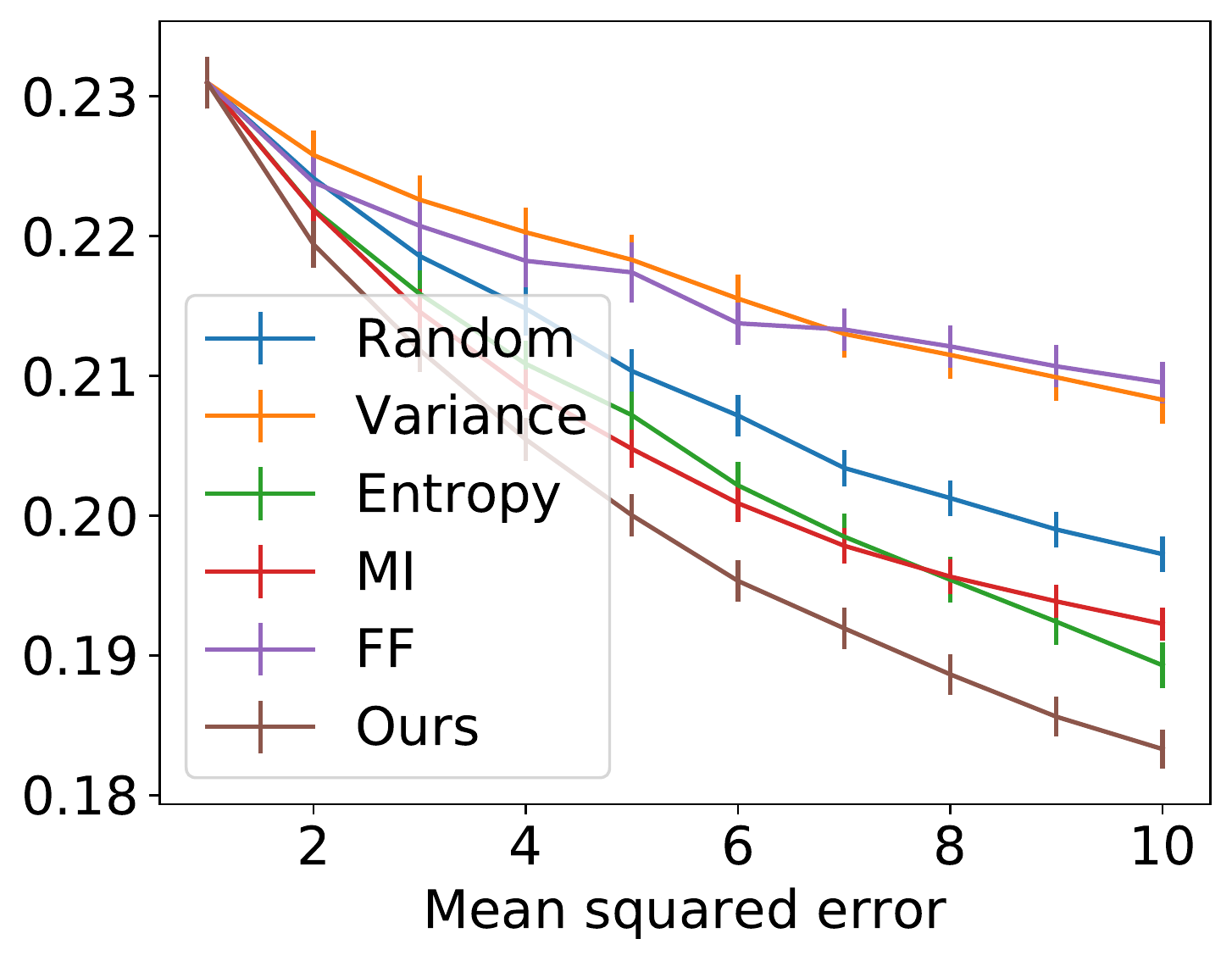}&
  \includegraphics[height=11em]{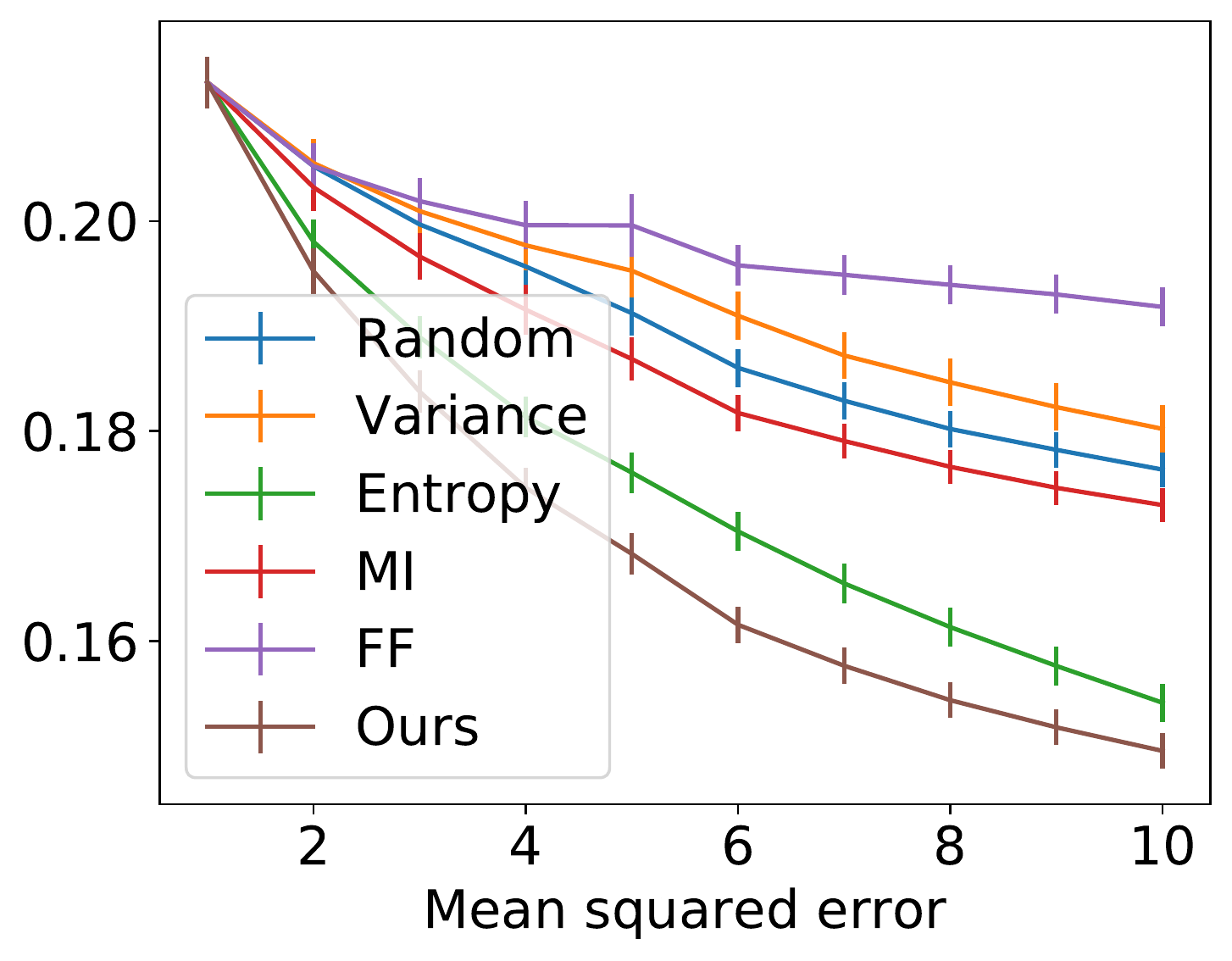}\\
  (a) pedal cycles & (b) two-wheeled & (c) car \\
  \\
  \includegraphics[height=11em]{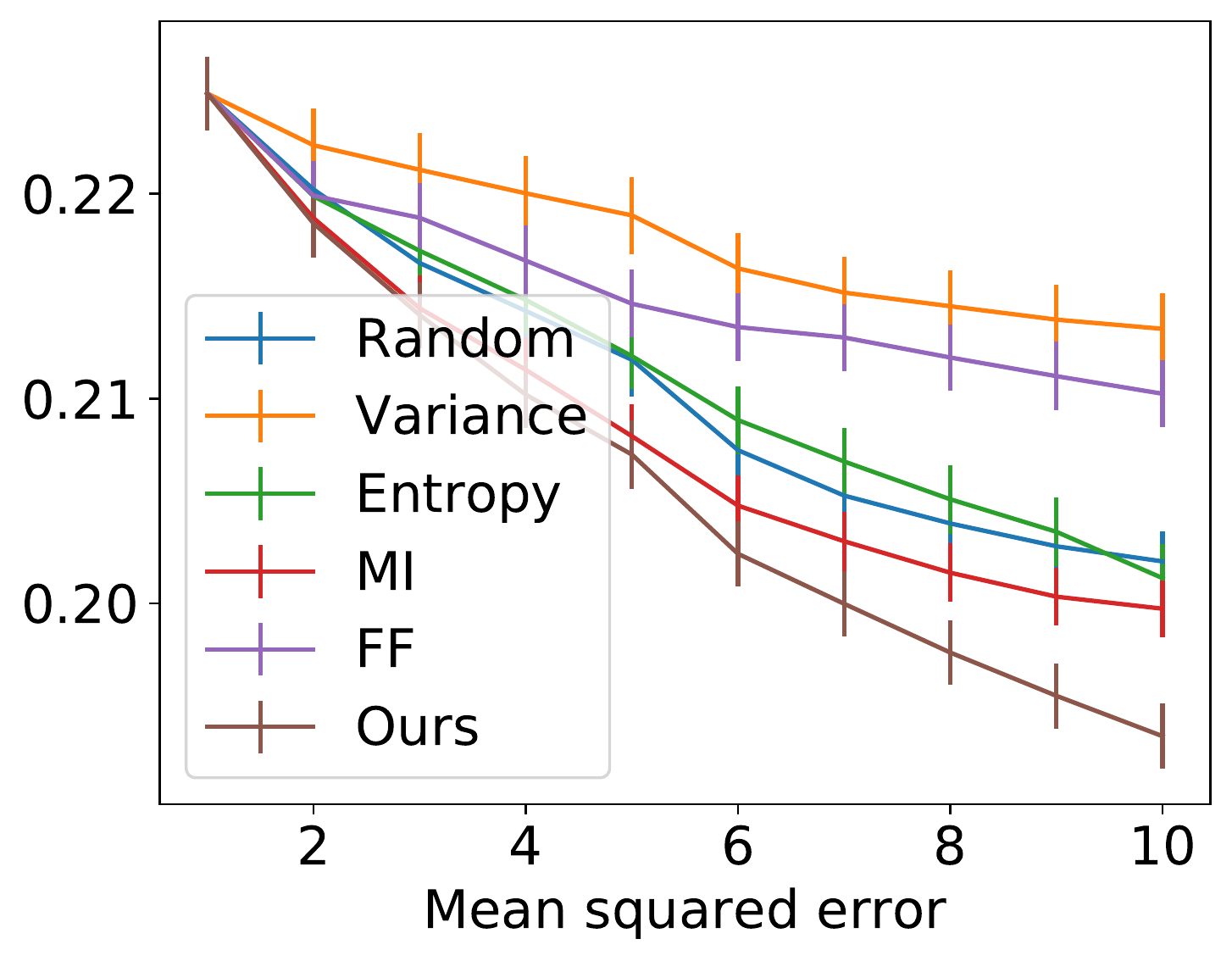}&
  \includegraphics[height=11em]{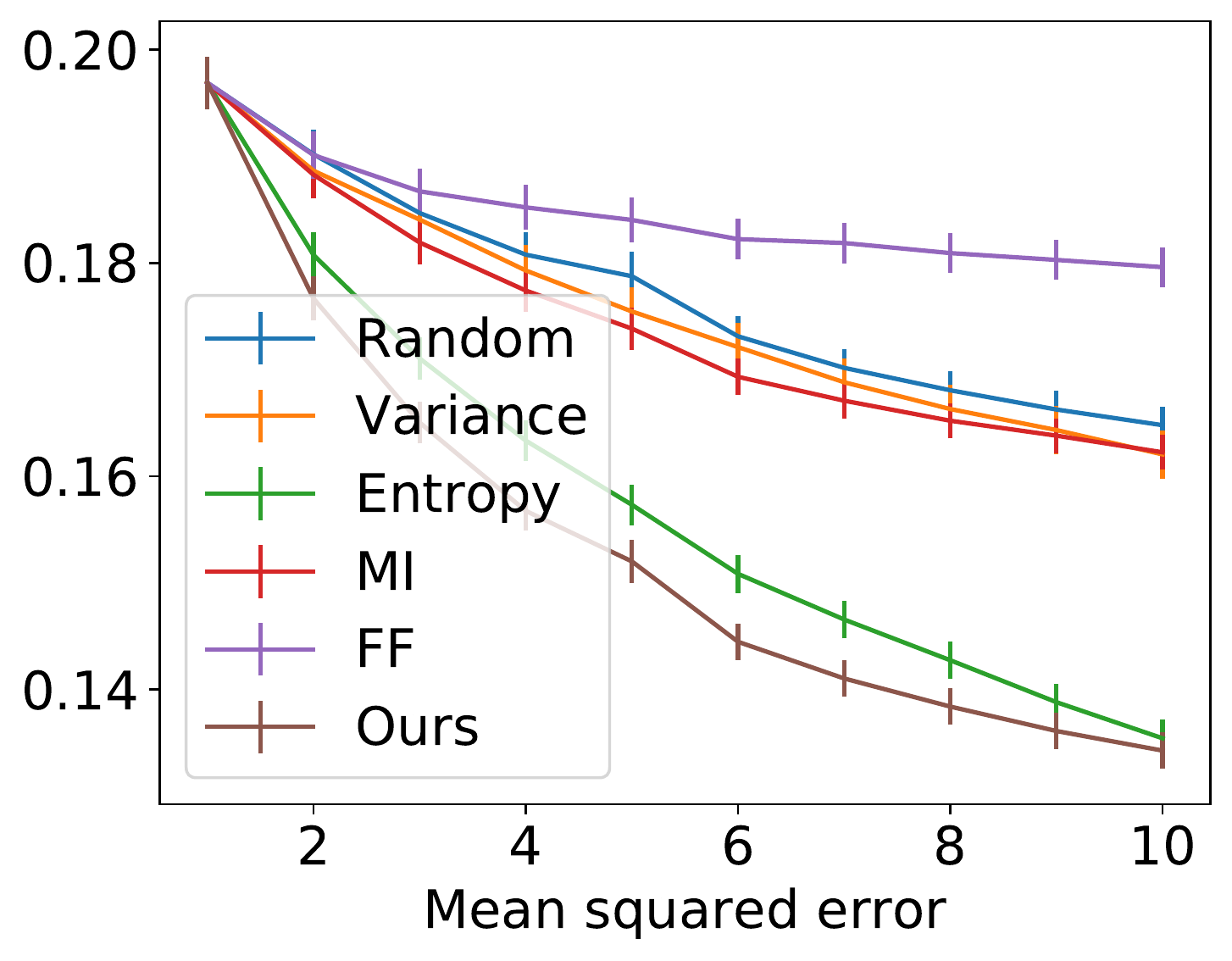}&
  \includegraphics[height=11em]{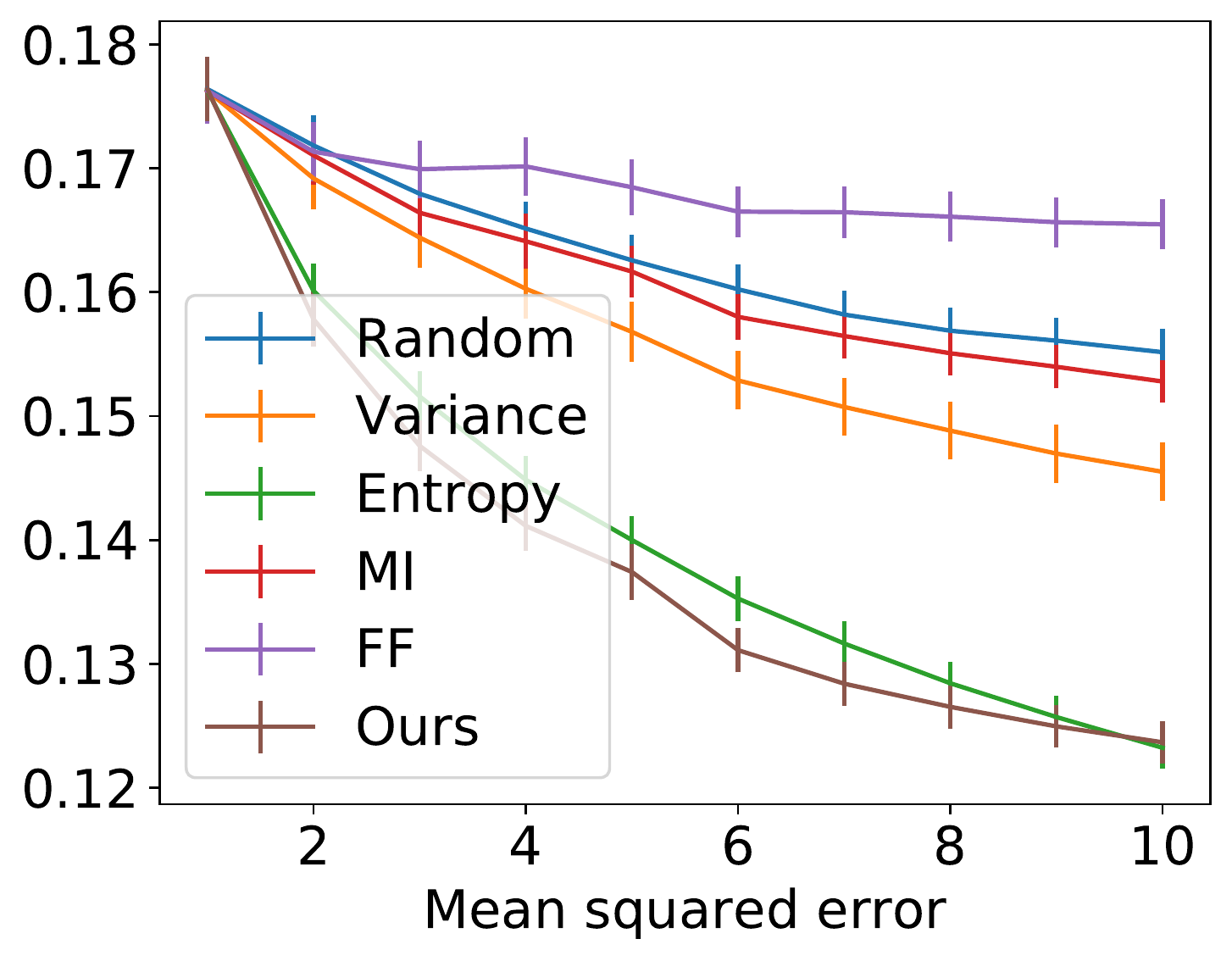}\\
  (d) buses and coaches & (e) LGVs & (f) 2-rigid \\\
  \\  \includegraphics[height=11em]{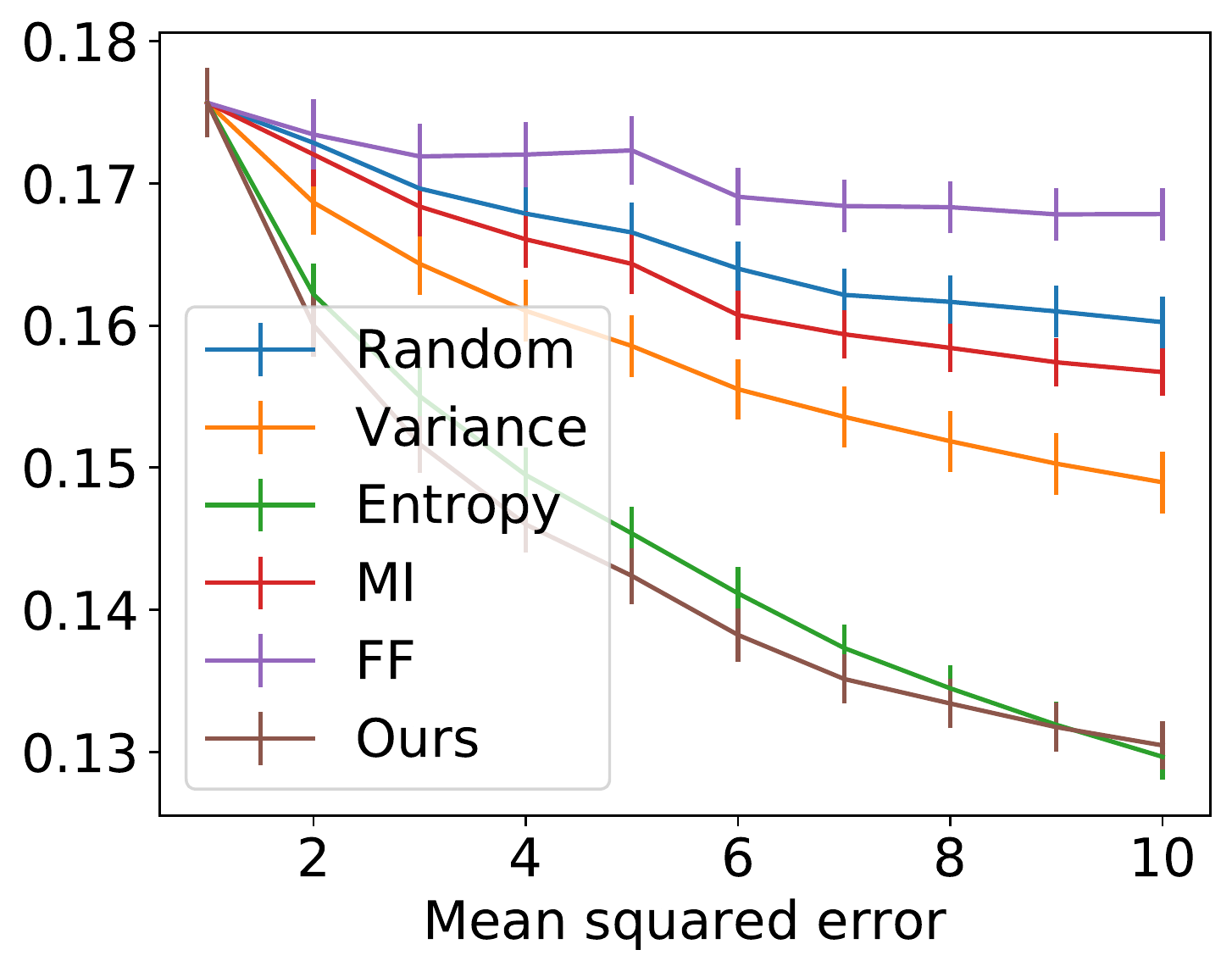}&
  \includegraphics[height=11em]{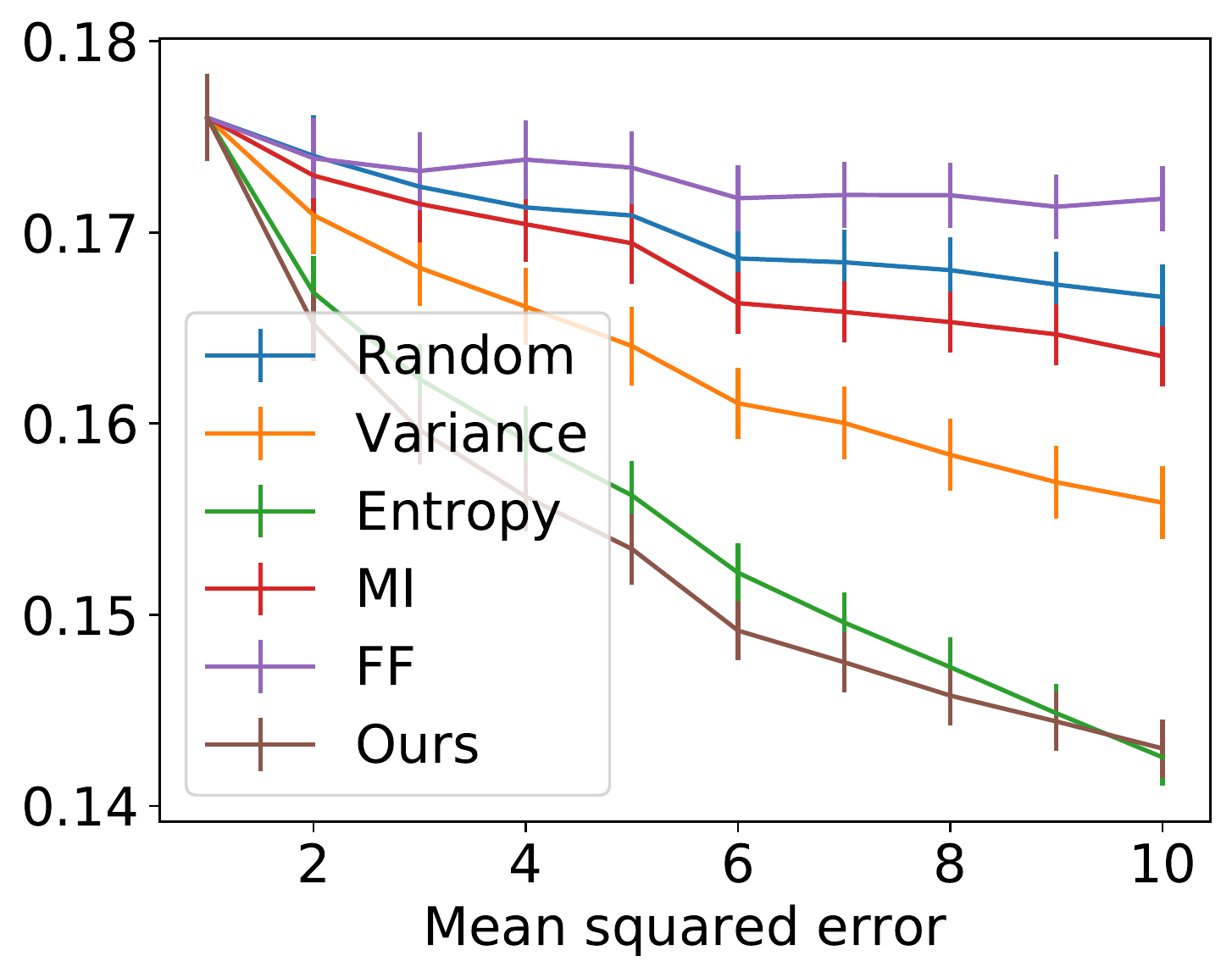}&
  \includegraphics[height=11em]{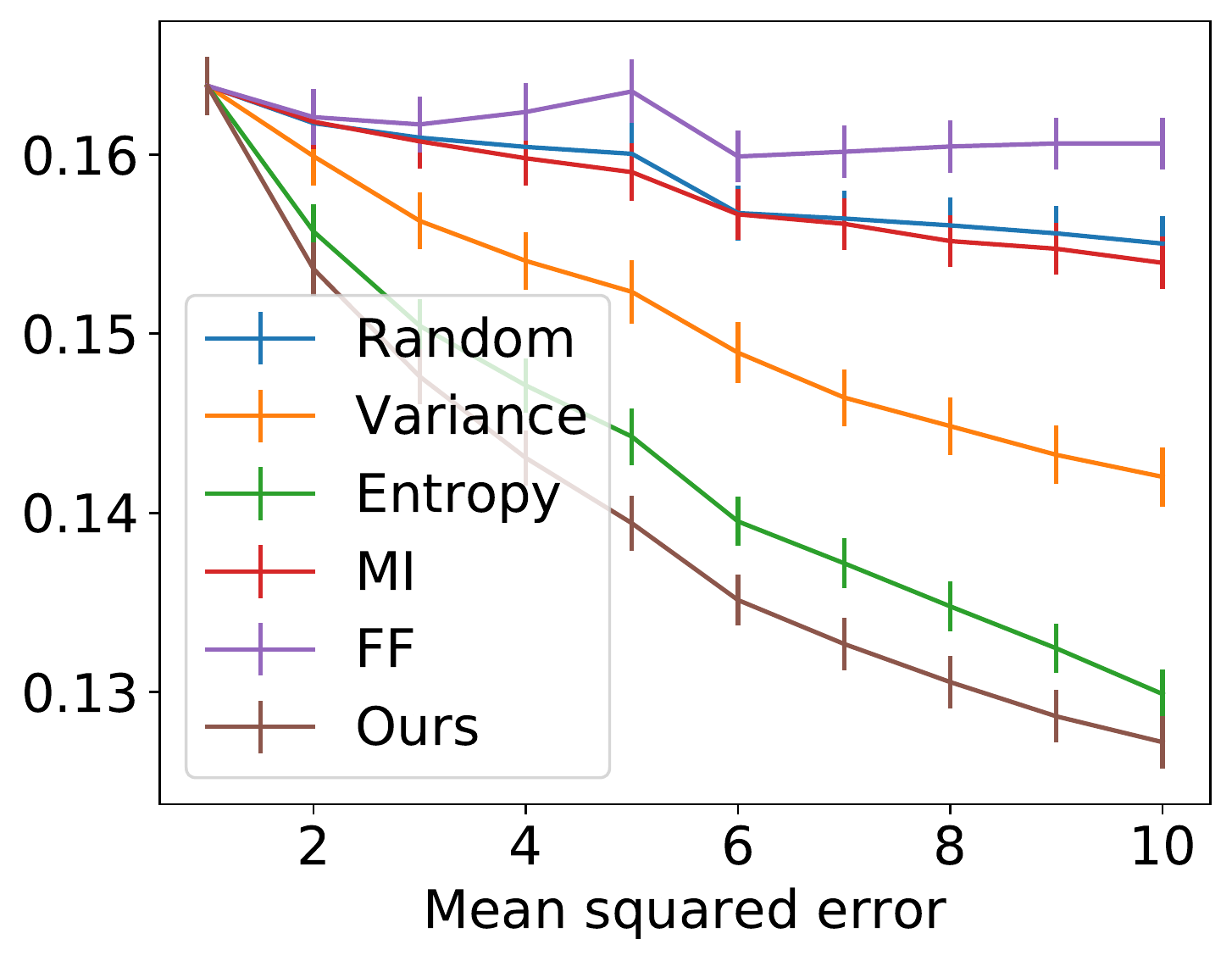}\\
  (g) 3-rigid & (h) 4-rigid & (i) 3-articulated \\
  \\
  \includegraphics[height=11em]{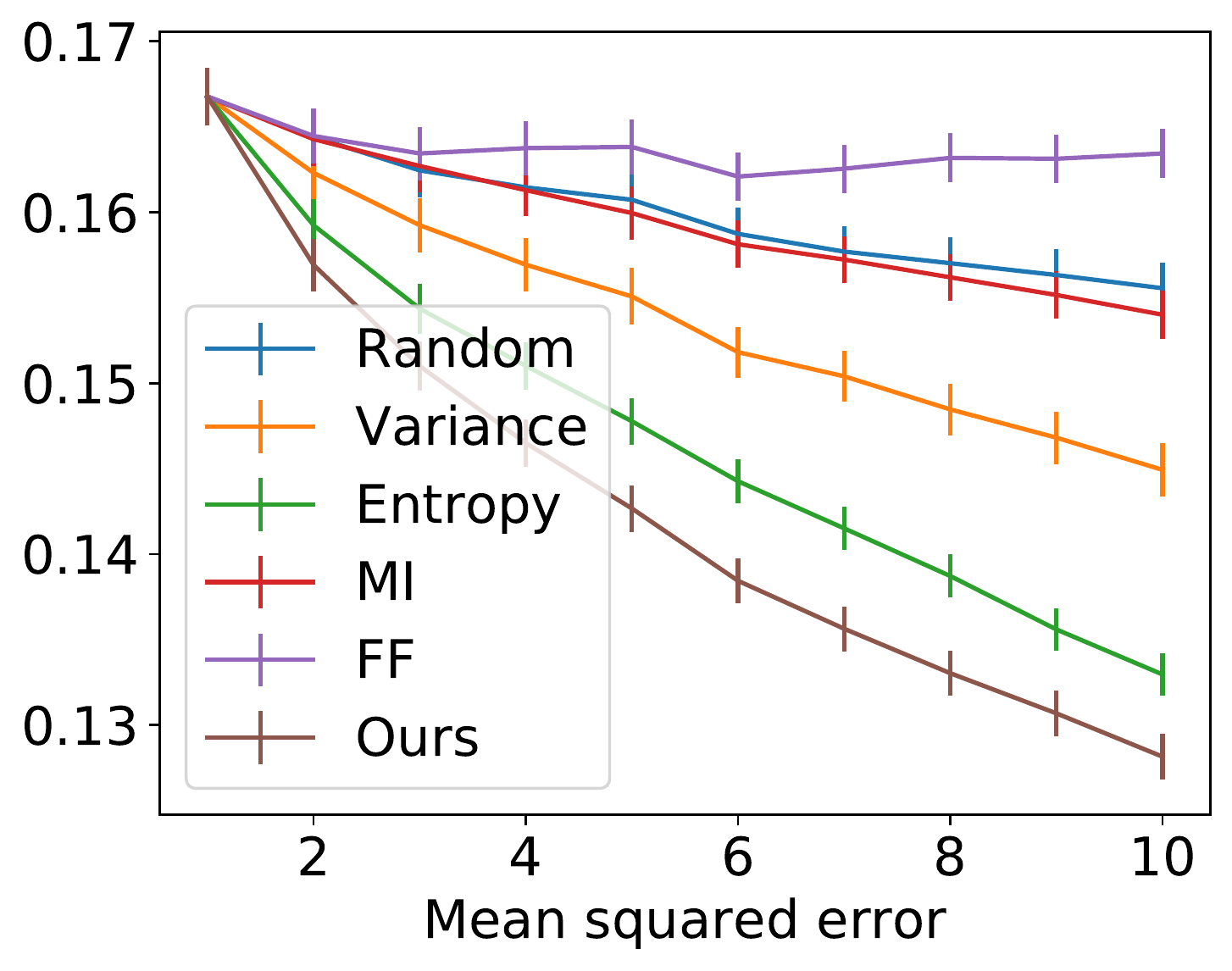}&
  \includegraphics[height=11em]{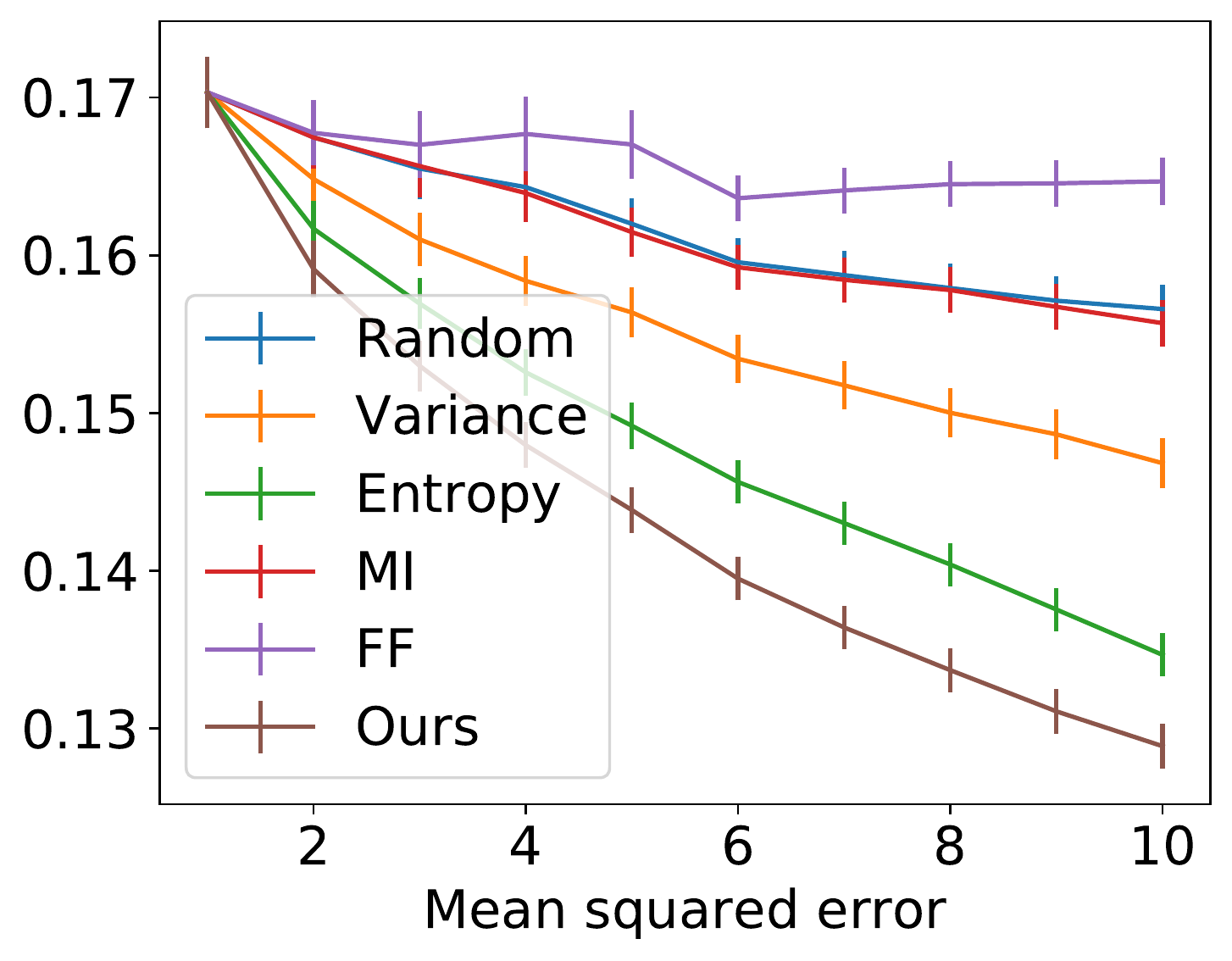}&
  \\
  (j) 5-articulated&
  (k) 6-articulated&
  \\
  \end{tabular}}
  \caption{Test mean squared error (vertical axis) with different numbers of observations (horizontal axis) for each AADF task. The bar shows the standard error.}
  \label{fig:results}  
\end{figure}

Figure~\ref{fig:results} shows the mean squared error and its standard error
with different numbers of observations for each AADF prediction task.
All the methods decreased the error as the number of observed nodes increased,
which implied that the response prediction model improved the performance by observing node responses.
The proposed method achieved the lowest mean squared error in most cases.
This result indicates that training node selection models
by directly maximizing test reduction errors using reinforcement learning is effective. 
The Entropy method achieved the second lowest error.
Even if we observed a node has high entropy,
the test errors for other nodes might not decrease
when the observed node has little relationships with other nodes.
On the other hand, the proposed method selects nodes so as to maximize the test error reduction
by considering the information of all nodes by GCNs.
The MI and Variance methods did not perform well.
This result implies that the estimation of variance and entropy based on dropout is difficult in this task.
Since the FF method cannot use the information of other nodes, it performed badly.
The computational time for training node selection and response prediction models with the proposed method
was 3.7 and 44.5 hours, respectively, by computers with 2.10-GHz Xeon Gold 6130 CPU.  

\section{Conclusion}
\label{sec:conclusion}

We proposed an active learning method for meta-learning on node response prediction tasks in graph data.
With the proposed method, graph convolutional neural networks are used for
calculating scores to select nodes to observe,
where attributes, observed responses, and masks indicating observed nodes are taken as input.
By using the graph convolutional neural networks,
we can select nodes for unseen graphs with unseen response variables
depending on a few observed responses and the graph structure.
We demonstrated that the proposed method achieved higher performance
for road congestion prediction tasks with fewer observed nodes
compared with existing active learning methods.
For future work, we want to apply the proposed method to tasks
other than node response prediction, such as link prediction and graph generation.
Although we used graph convolutional neural networks,
we plan to use other meta-learning models for meta-active learning.

\bibliography{neurips2020graph}
\bibliographystyle{abbrv}

\end{document}